\documentclass[a4paper, 10pt, conference]{ieeeconf}      
\usepackage[utf8x]{inputenc}
\usepackage{tabularx}
\usepackage{graphicx}  
\usepackage{float}  
\usepackage{caption}
\usepackage{balance}
\IEEEoverridecommandlockouts                              
\usepackage{graphics} 
\usepackage{epsfig} 
\usepackage{mathptmx} 
\usepackage{times} 
\usepackage{amsmath} 
\usepackage{cuted}
\usepackage{amssymb}  
\usepackage{comment}
\usepackage{svg}
\usepackage{authblk}

\title{\LARGE \bf
Attention based Occlusion Removal for Hybrid Telepresence Systems}

\author{\Large Surabhi Gupta \qquad Ashwath Shetty \qquad Avinash Sharma \\ 
\\
\large International Institute of Information Technology Hyderabad \\ \textit{ \{surabhi.gupta, ashwath.shetty\}@research.iiit.ac.in, asharma@iiit.ac.in}}

\begin{document}

%
\maketitle
\begin{strip}
    \centering
    \includegraphics[width=\linewidth]{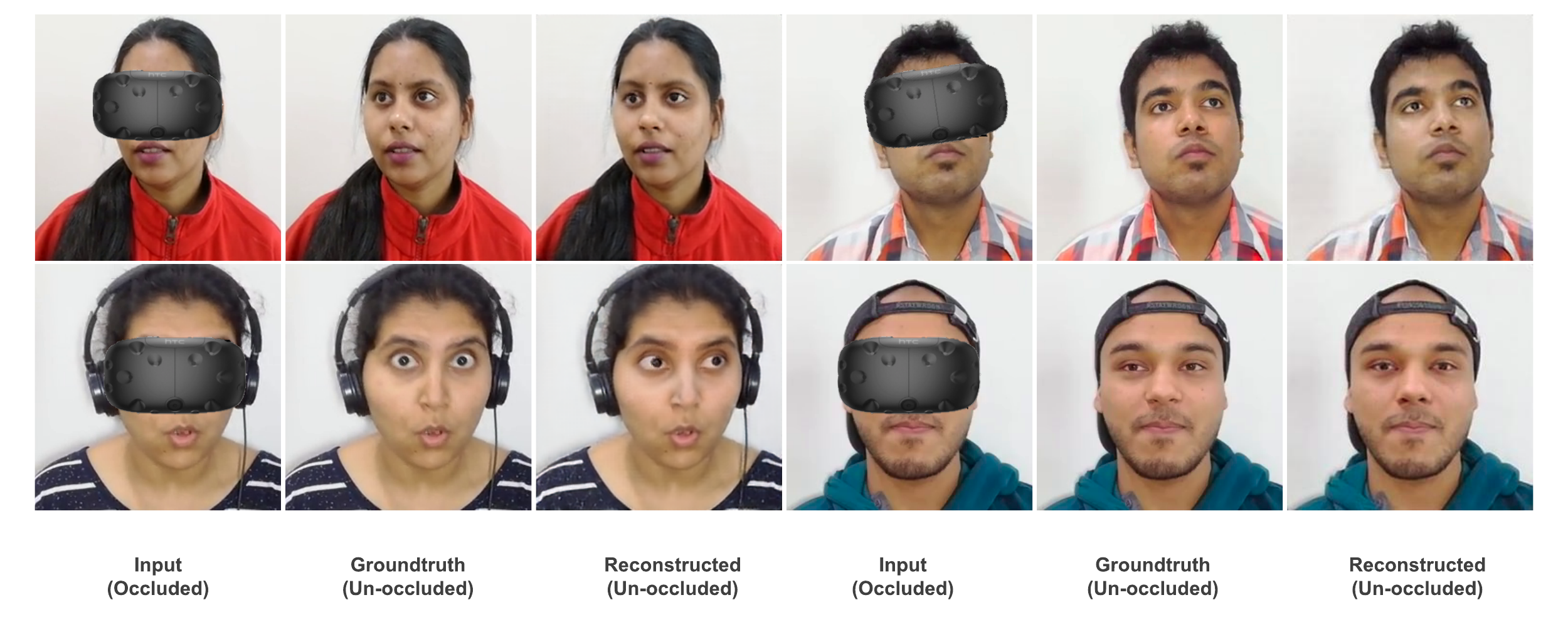}
    \captionof{figure}{Our proposed approach can reconstruct high-quality unoccluded image from a given occluded face image as input.}
    \label{fig:teaser}
\end{strip}
\setcounter{figure}{1}  

\begin{abstract}
Traditionally, video conferencing is a widely adopted solution for telecommunication, but a lack of immersiveness comes inherently due to the 2D nature of facial representation. The integration of Virtual Reality (VR) in a communication/telepresence system through Head Mounted Displays (HMDs) promises to provide users a much better immersive experience. 
However, HMDs cause hindrance by blocking the facial appearance and expressions of the user. To overcome these issues, we propose a novel attention-enabled encoder-decoder architecture for HMD de-occlusion. We also propose to train our person-specific model using short videos (1-2 minutes) of the user, captured in varying appearances, and demonstrated generalization to unseen poses and appearances of the user. We report superior qualitative and quantitative results over state-of-the-art methods. We also present applications of this approach to hybrid video teleconferencing using existing animation and 3D face reconstruction pipelines.\\

\end{abstract}


\section{INTRODUCTION}

Globalization has led to an acute need for tele-interactions for effective communication that has been further boosted due to the current pandemic situation across the world. Traditionally, video conferencing is a widely adopted solution for telecommunication, but a lack of immersiveness comes inherently due to the 2D nature of facial representation. The integration of Virtual Reality (VR) in a communication/telepresence system through Head Mounted Displays (HMDs) promises to provide users a much better immersive experience. 
Nevertheless, HMDs significantly occlude the user's face, hindering facial appearance capture, including gaze and expressions.  Therefore, HMD removal in images is vital for enabling an improved user experience. 

Traditionally, Analysis-by-Synthesis techniques for HMD de-occlusion proposed in the literature  \cite{19} animate parametric face models \cite{20} using features extracted from an HMD occluded input image. However, such models often generate overly smooth geometrical details and compromise the realism of facial appearance. On the other hand, recent facial avatar-based models \cite{10} achieves photorealistic results for HMD de-occlusion. However, avatar modeling methods require a large amount of calibrated multi-view data of a single user in different poses and expressions for avatar creation (e.g., \cite{10} uses setup consisting of 40 machine vision cameras capable of synchronously capturing 5120×3840 images at 30 frames per second). Thus, such avatar creation is non-trivial and challenging to scale up for a large user base. Additionally, it is a one-time process for each user. Therefore, such parametric model or avatar-based techniques have a significant limitation: they lack the user's actual appearance and background during the interaction (i.e., unable to model the everyday appearance of the user), thereby hindering the user experience.

\begin{figure}[h]
    \includegraphics[width=\linewidth]{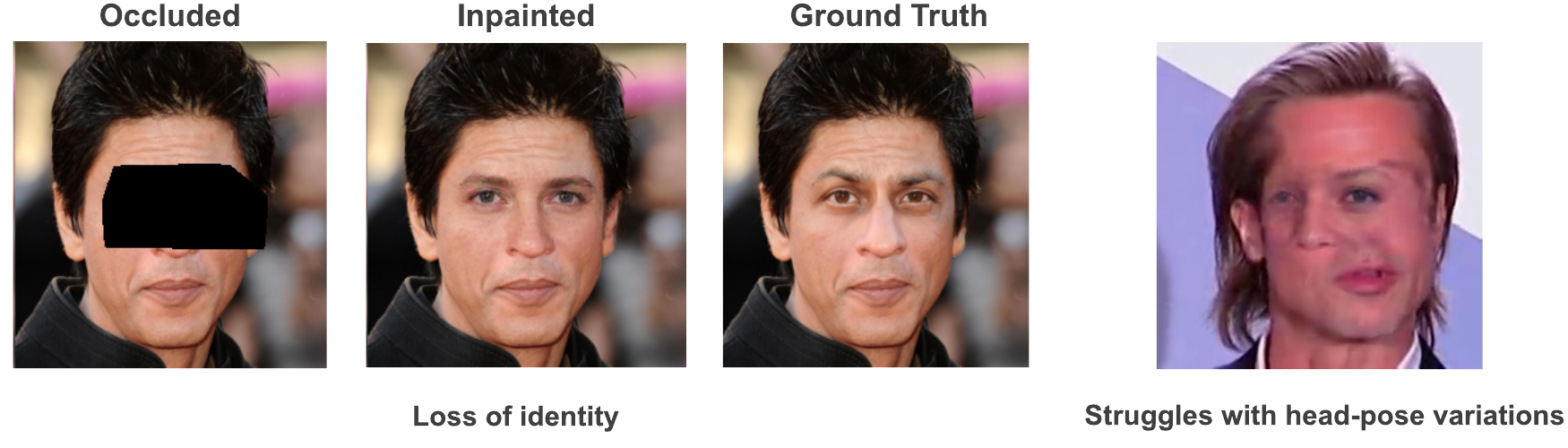}
    \caption{Failure cases of LaFin~\cite{1}}
    \label{fig:failure}
\end{figure}

HMD de-occlusion problem can also be posed as a face completion/inpainting problem. 
Existing face completion methods in literature attempt to learn a single inpainting network over a large training population~\cite{1, 13}, hoping for good generalization on unseen face images. However, these methods frequently suffer from issues like loss of identity and fail to generalize with even minor variations in head pose, as shown in Figure~\ref{fig:failure}. Another set of methods in~\cite{2,3} uses a reference image along with the occluded area to fill a masked region and preserve identity. However, as shown in~\cite{2}, their work fails to generalize with non-frontal head-poses and \cite{3} requires additional information (depth and mask) and do not generate the entire face (with hairs, ears, and background), which hinders user experience.

The primary research challenge with the face completion/inpainting task comes from it's ill-posed nature as a significant part of the face is occluded by HMD. Learning a single common face inpainting/de-occlusion network with the capability to hallucinate diverse expressions in varying appearances and head poses across a large set of human faces is difficult to achieve. It is due to the broad space of facial geometry and appearance as well as the highly subjective way of articulating expressions/emotions across individuals~\cite{survey}. Additionally, in the context of VR teleconferencing applications, the desired solution should be scalable, requiring minimal efforts and hardware at the user's end to capture their everyday appearance. 
An additional desired characteristic might be regarding integration ability in a hybrid VR teleconferencing setup where users with only video capability should also be able to participate as in regular video conferencing. 

To overcome these challenges, we propose to tackle this problem in a person-specific setting where we aim to train a dedicated model for each user to learn user-specific appearance, head-pose, gaze, and facial expression traits. 
The input to our method is a video frame with HMD occluded face, and our model generates a de-occluded plausible face by removing the HMD occlusion.  To achieve this, we introduce a novel attention enabled encoder-decoder architecture and a novel training strategy to train our person-specific model using short videos (1-2 minutes) of the user, capturing varying appearance of the user/background and more specifically variety of head poses and facial expressions without HMD occlusion. As part of our training strategy, we first train the encoder-decoder module (sans attention) on large face image datasets to learn generic face appearance features. Subsequently, we finetune it on unoccluded user videos. Finally, we train/finetune our full model (encoder-decoder with attention) on the same training data with synthetic HMD masks. It allows our model to learn the person-specific facial geometry and expression traits and help generate occluded areas with varying appearances, poses, and expressions.  
Our attention module allows the network to preserve high-frequency appearance/background details (like hairs, wall texture, etc.) from the unoccluded part of the input HMD occluded image while generating the plausible facial appearance for the occluded part. Our novel mask-loss function helps the model to emphasize the occluded region. Figure 1 shows high-quality de-occlusion achieved by our method. It is important to note that learning a person-specific model is not equivalent to overfitting on a specific user as there is a significant change in user appearance, background, and lighting across sessions. Similar person-specific model learning has been successfully explored in this context (e.g., ~\cite{10}) as well as another related context of egocentric frontal face recovery~\cite{6}. 
We conduct a thorough empirical evaluation and report superior qualitative and quantitative results of our face de-occlusion method w.r.t. SOTA methods. 
Additionally, we also demonstrate the broader applicability of the proposed HMD de-occlusion method with two use cases. First, it can be used to build hybrid VR systems by integrating it with \cite{one-shot,5} kind of video-driven face animation solutions. As a second use case, we show how we can integrate our method with a 3D face reconstruction pipeline to generate 3D face video for VR teleconferencing system. 
To summarise, we make the following contributions:

\begin{enumerate}
  \item We present a deep learning framework for person-specific HMD occlusion removal. Our method does not rely on high-end hardware (e.g., HMD with gaze tracking) and calibrated data from the user (for avatar creation).
  \item We introduced a novel attention module knitted with the encoder-decoder architecture that has the ability to use background and appearance details from the input image and allows the model to focus on inpainting the occluded region with plausible details.
  \item We collected a small dataset of multiple users in different appearance/facial expression/head-pose and lighting variations. We intend to release this for the academic community. 
  \item We also present applications of our model where it can be integrated with neural animation models \cite{one-shot,5} to animate an occluded video and can also be used to recover 3D face from occluded input.
\end{enumerate} 

\section{RELATED WORK}
Our approach is an inpainting method that learns to fill in user-specific details. It is related to traditional inpainting methods and recent approaches that use a reference image to fill an occluded region faithfully. In the following, we discuss the most relavent literature in detail.

\subsection{Person Specific Models}
\label{sec:Person_Specific}
Recent deep learning-powered advances in vision and graphics have led to the rise of personalized models that are animated/rendered using deep learning. \cite{10}  learns auto-encoder network to predict view conditioned texture and mesh geometry from HMD occluded input. \cite{11} trains a dynamic neural radiance field model on a short video (2-3 min) of the person with different expressions and poses. Also closely related to our work is~\cite{6}. They use a video to video GAN \cite{15}, which takes in egocentric frames of a person and generates the corresponding frontal view. These methods show the ability of person-specific models to capture high-frequency details. However, they require calibrated multi-view data from the user, which adds additional hardware constraint, and animating these models is also expensive. Our approach requires the user to send a short uncalibrated video session for training and is considerably lightweight compared to avatar-based methods.
\begin{figure*}
    \includegraphics[width=\textwidth]{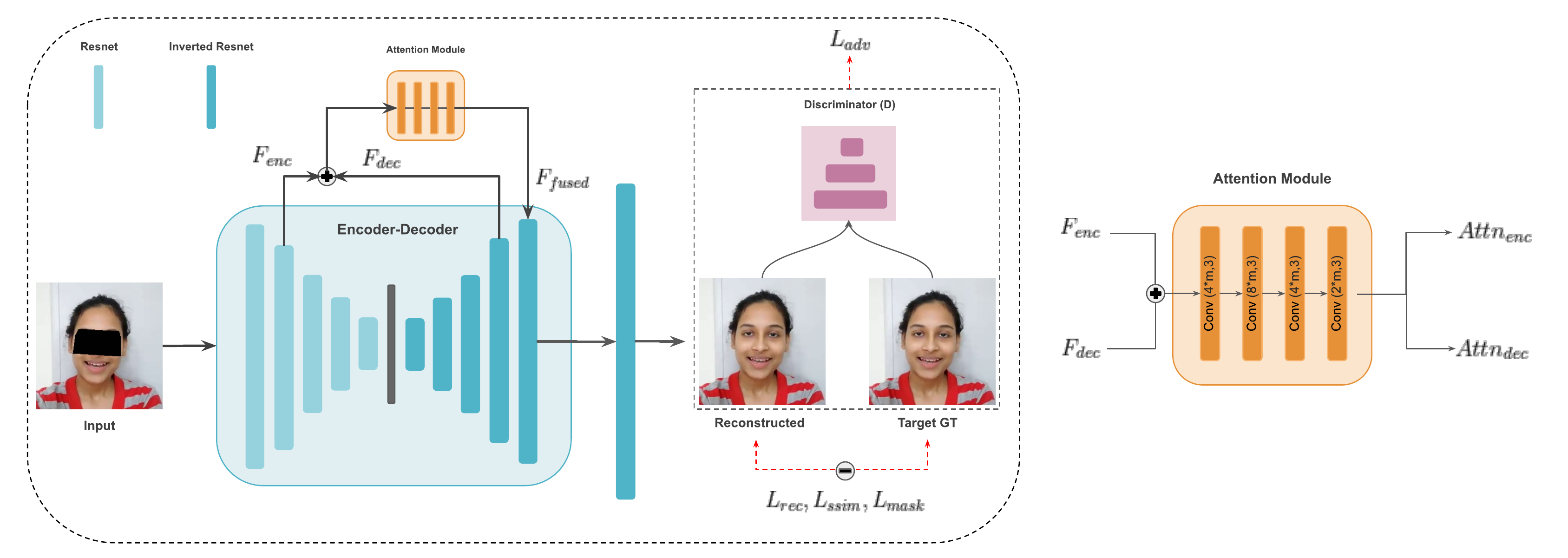}
    \caption{Detailed architecture diagram of our proposed network.}
    \vspace{-1.5em}
    \label{fig:model}
\end{figure*}
\subsection{Image Inpainting Methods}
\label{sec:Image_Inpainting}
Image inpainting describes the task of filling missing image regions with realistic content. Recent work \cite{13,14} and \cite{1} train a conditional GAN \cite{16} on a face dataset as a solution to this problem. These methods show impressive generalization to examples with frontal head pose and arbitrary occlusion. However, they are biased to their training distribution, do not generalize well to even slightly non-frontal head poses, and suffer from a loss of identity. As we train our method only on images of the same person with considerable variations in head poses and expressions, we can overcome the challenges of identity loss and manage to generalize to various head poses.

\subsection{HMD Removal}
\label{sec:hmd_removal}
 Recent works like \cite{3} and \cite{2} propose an image-based approach to HMD de-occlusion. They try to solve the identity problem mentioned above by using a reference image to guide the inpainting procedure and learn a general model for the task. However~\cite{2} fails to work well with cases of pose variations between the reference and occluded image and suffers from loss of identity in these cases. Also, \cite{3} requires additional depth data and train and evaluate on synthetic data, which may not work or be available in a real-world teleconferencing scenario. We train and evaluate our model on data of real-world conversations and scenarios and show our model's ability to generalize to unseen appearances.

\addtolength{\textheight}{-1cm}   


\section{METHODOLOGY}

\subsection{Overview}
\label{sec:overview}
The primary focus of our work is to learn a personalized model for face de-occlusion, particularly as an application in VR teleconferencing, where the face is partially occluded due to HMD. To tackle this, we formulate the face de-occlusion problem as an image inpainting task. Given an occluded face image as an input  $X_{occ}$, our network aims to hallucinate the missing region with plausible and perceptually consistent facial details in order to reconstruct the generated unoccluded image, $X_{rec}$ against the ground truth unoccluded image, $X_{gt}$. 

Inspired by the autoencoder architecture proposed in~\cite{7}, we use an novel attention enabled encoder-decoder framework with generative capabilities that learns to reconstruct high-fidelity unoccluded faces from HMD occluded input images. Additionally, we also propose a novel mask-based loss function and a novel training strategy to learn our model. 
Figure~\ref{fig:model} shows the outline of our proposed architecture.

\subsection{Proposed Architecture}
\label{sec:method}
\subsubsection{Encoder-Decoder Module}
\label{sec:recon-module}
Our encoder-decoder module comprises a stack of ResNet and inverted ResNet blocks. Each ResNet block consists of a set of convolutions with residual connections. For the inverted ResNet block, the first convolution in the ResNet block is replaced by a $4 \times 4$ deconv layer.  We also provide the additional generative capability to the network using an adversarial loss. The encoder learns a 99-dimensional feature representation of the input image. This bottleneck representation is subsequently fed to the decoder network to reconstruct the target image.

\subsubsection{Attention Module}
\label{sec:atten-module}
Inspired from existing literature on attention-based learning strategies as proposed in \cite{attn,ashishACCV}, we append our encoder-decoder module with an attention module.
We perform spatial attention by taking the encoder output from the second layer, $F_{enc}$ of spatial dimension $64 \times 64$ and perform a channel-wise concatenation with the corresponding decoder layer output, $F_{dec}$ (i.e., with same spatial dimension) and feed it to our attention module which subsequently generates attention maps of same dimension as shown in Figure.~\ref{fig:model}(A). We then decouple these attention maps and use them for a weighted fusion of respective feature maps (i.e., $F_{enc}$ and $F_{dec}$). The fused feature maps are subsequently fed downstream to convolution layers to reconstruct the de-occluded face image. 

Such attention-based spatial feature map fusion allows our network to preserve high-frequency appearance/background details  (like hairs,  wall texture,  etc.)  from the visible part of the input image while generating a plausible facial appearance for the occluded part. 

These attention maps can be learned using fully convolutional networks. 
As shown in Figure.~\ref{fig:model}(B), the attention module consists of ${Conv(4∗m,3)}$, $Conv(4∗m,3)$, $Conv(8∗m,3)$ and $Conv(2∗m,3)$, where $m$ denotes the base number of filters and $Conv(m,k)$ denotes a convolutional layer with output number of channels $m$ and kernel size $k$.
The final output with $2*m$ channels is then split into two, $Attn_{enc}$ and $Attn_{dec}$, each of $m$ channels and spatial dimension of $64 \times 64$. This acts as a attention mask for the inputs, $F_{rec}$ and $F_{dec}$ which is then fused again using a channel-wise summation according to Equation~\ref{eq:attn}.
\begin{equation}
F_{fused} = F_{enc}*Attn_{enc} + F_{dec}*Attn_{dec}
\label{eq:attn}
\end{equation}

\subsubsection{Loss Function}
\label{sec:loss-module}
We employ a combination of four different loss functions as our training objective.
First, in order to penalize reconstruction errors, we use pixel-based $L1$ loss.
\begin{equation}
L_{rec} =  \lVert X_{gt}-X_{rec} \rVert_{1}
\label{eq:l1-loss}
\end{equation}
However, using only the $L1$ reconstruction loss produces blurry outputs. To overcome this, we add a discriminator, $D$ in the architecture to compute the adversarial loss. This adversarial loss term forces the encoder-decoder to reconstruct high-fidelity outputs by sharpening the blurred images. For $D$, we adopt the architecture of the DCGAN discriminator\cite{radford2016unsupervised}.

\begin{equation}
L_{adv} = log(D(X_{gt}))+log(1-D(X_{rec}))
\label{eq:adv-loss}
\end{equation}
 We also use SSIM based structural similarity loss (as defined in \cite{7}) that helps to improve the alignment of high-frequency image elements to stabilize the adversarial training.
\begin{equation}
L_{ssim} = SSIM(X_{rec},X_{gt}) 
\label{eq:ssim-loss}
\end{equation}
To further improve the quality of reconstruction in the HMD occluded area of the generated image, we propose a novel \textit{mask-based loss}. Here, we use the binary mask image as an additional supervision to the network along with input image while training. 
Minimizing this loss helps the model to emphasize more on quality of reconstruction in the masked region. 
This also helps to mitigate the blinking artifacts around the eye region for stable reconstructions.
We formulate the mask-based loss function as:
\begin{equation}
L_{mask} = \lVert I_{mask}\odot I_{gt} - I_{mask}\odot I_{rec} \rVert_{1}
\label{eq:mask-loss}
\end{equation}
where, $I_{mask}$ refers to single channel binary mask image where white pixels (1) correspond to occluded region and black pixels (0) correspond to the remaining unoccluded region and $\odot$ is element-wise multiplication. 
Thus, the final training objective loss function can be written as,
\begin{equation}
Loss = \lambda_{rec}*L_{rec}+\lambda_{adv}*L_{adv}+\lambda_{ssim}*L_{ssim}+\lambda_{mask}*L_{mask}
\end{equation}
where, $\lambda_{rec}, \lambda_{adv}, \lambda_{ssim}$ and $\lambda_{mask}$ are the corresponding weight parameters for each loss term.
\subsection{Our Training Strategy}
\label{sec:training-strategy}
For the task of learning a person-specific model, we adopt a two-step training process, where the first step and second step are trained on unoccluded and occluded face images of the person, respectively. In the first step, we freeze the attention module and only train the encoder-decoder to reconstruct the unoccluded images. Here we start with unsupervised training of only encoder-decoder on publicly available state-of-the-art face datasets such as VGGFace~\cite{9} and AffectNet~\cite{8} to leverage the inherent knowledge about the face structure. This enables the model to grasp knowledge about the basic facial structure and features such as eyes, nose, mouth, etc. This step is common for all users, and later we perform finetuning on user's images with a wide range of pose, expression, and appearance variations. This step helps the model to learn the exact geometry of the user's face. 

In the second step, we unfreeze our attention module and train the entire architecture on occluded images. This can be considered as a self-supervised learning approach, where the input is an occluded image, and the target image is its corresponding unoccluded image. We, therefore, minimize the loss between the reconstructed face image and the unoccluded ground truth image. This enables the attention module to learn to retain the high-frequency details from the visible part of the occluded input image while performing a soft fusion with the generated image. Thus, our two-step training strategy yields superior de-occlusion with no explicit boundaries between occluded and visible regions of the reconstructed image.  

\begin{figure*}
\centering
   \includegraphics[width=\textwidth]{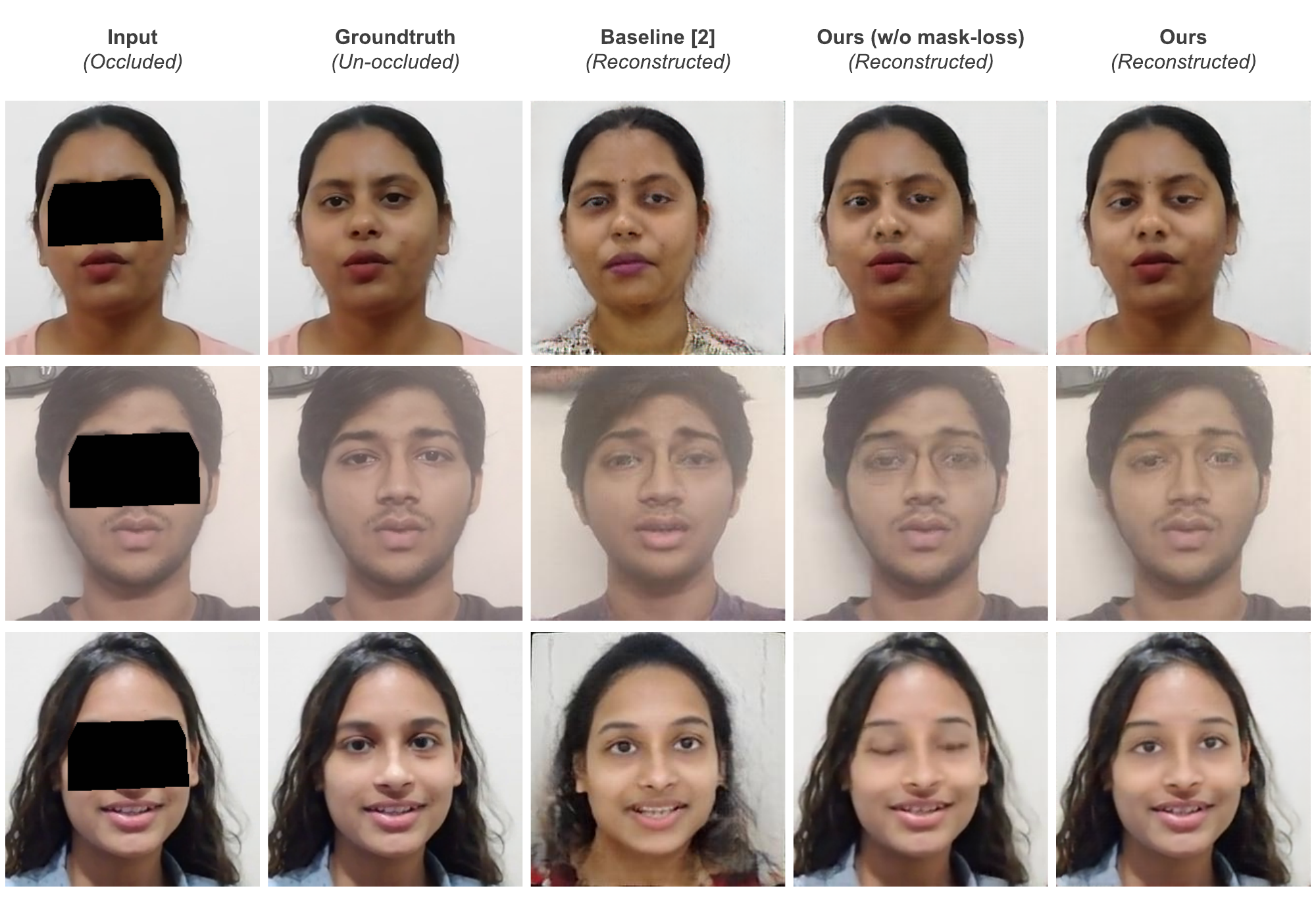}
    \caption{Comparison with baseline and variants of our method.}
    \vspace{-1.0em}
    \label{fig:arch comparison}
\end{figure*}

\section{EXPERIMENTS AND RESULTS}

\subsubsection{Dataset}
\label{sec:dataset}

Our method uses short monocular RGB video sequences. Hence, we captured various human subjects in different appearances at a 1280x720 pixels resolution with 30 frames per second frame rate from a mobile phone camera. The images are cropped and scaled to 256x256. These sequences have a length of around 2 min, i.e., approx—3600 frames per session. We use mutually exclusive sets of these video sequences from the same subject in different appearances to train and evaluate our user-specific model. The subjects were asked to engage in day-to-day conversation and demonstrate variations in head poses.

To create a synthetic mask that simulates an HMD for each video frame, we generated a binary mask guided by the facial landmarks and placed it over the eye region to occlude the face. 

\subsubsection{Implementation Details}
\label{sec:implementation}
For step 1 of training strategy (see Section~\ref{sec:training-strategy}), we train the encoder-decoder architecture on unoccluded images with three loss functions (i.e., Equations~\ref{eq:l1-loss},\ref{eq:adv-loss},\ref{eq:ssim-loss}) in a stage-wise manner with each loss term is being added in the training objective with every stage. We chose a batch size of 50 and an input resolution of 256x256. We train the network for 300, 100, and 300 epochs, respectively, for each loss function's incremental addition. For step 2, we similarly train the encoder-decoder architecture with an additional attention module with mask loss  (Equation~\ref{eq:mask-loss}) on occluded images for 300, 100, and 200 epochs, respectively, for each of the incremental addition of loss functions.
We use the Adam optimizer \cite{18} with a constant learning rate of 0.00002. We use $\lambda_{rec}=1, \lambda_{adv}=0.25, \lambda_{ssim}=60$ and $\lambda_{mask}=1$.

\subsubsection{Evaluation Protocols}
\label{sec:evaluation_protocols}
We choose SSIM (Structural Similarity Index), PSNR (Peak signal-to-noise ratio), and LPIPS (Learned Perceptual Image Patch Similarity) as our evaluation metrics for quantitative comparisons. For SSIM and PSNR, higher the value better the reconstruction quality, and for LPIPS, lower the value better the perceptual quality.

\subsubsection{Results}
\label{sec:results}
In this section, we demonstrate the superiority of our framework in terms of qualitative and quantitative results. For a baseline method, we train the architecture from \cite {7}, with the strategy mentioned in ~\ref{sec:training-strategy}. 
\begin{figure*}
\centering
    \includegraphics[width=\textwidth]{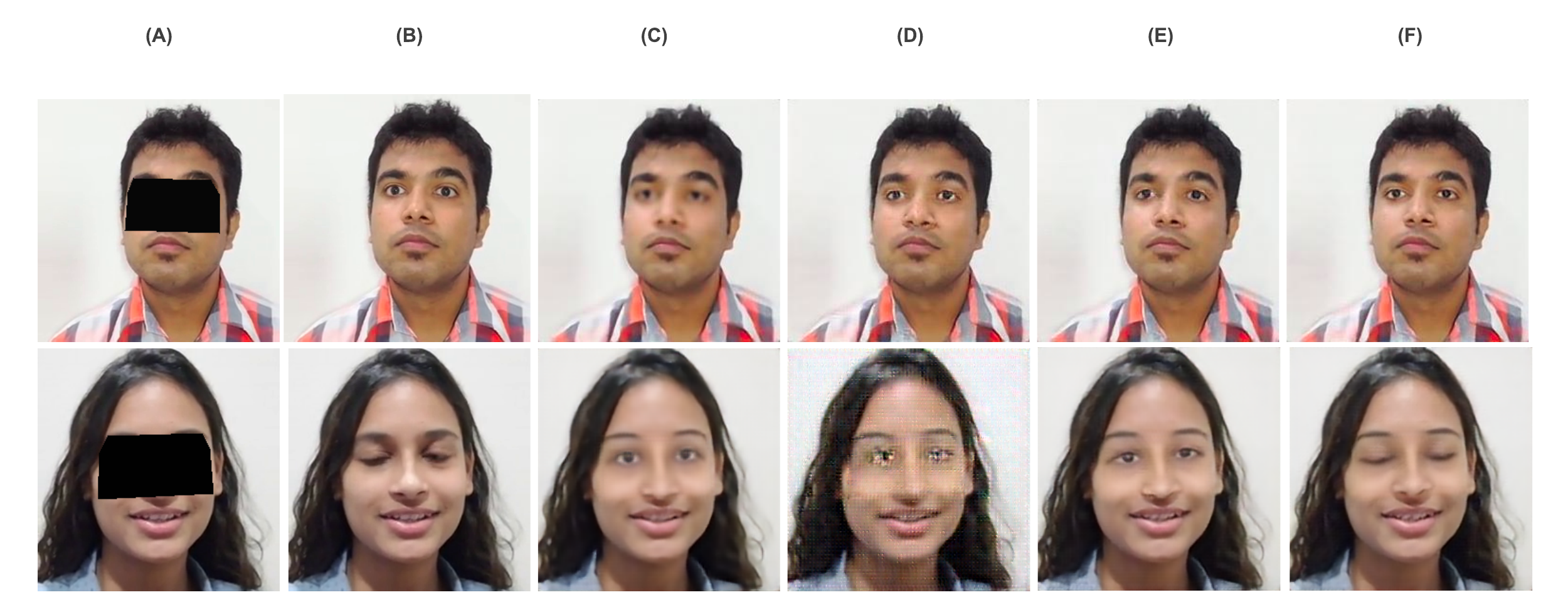}
    \caption{Qualitative comparison with different loss functions. From left to right: (A) Input, (B) GT, (C) with $L_{rec}$, (D) with  $L_{rec}$ and  $L_{adv}$, (E) with  $L_{rec}$,  $L_{adv}$ and  $L_{ssim}$, (F) Ours (with  $L_{rec}$,  $L_{adv}$,  $L_{ssim}$ and  $L_{mask}$).}
    \vspace{-0.5em}
    \label{fig:loss}
\end{figure*}

\begin{figure*}
\centering
    \includegraphics[width=\textwidth]{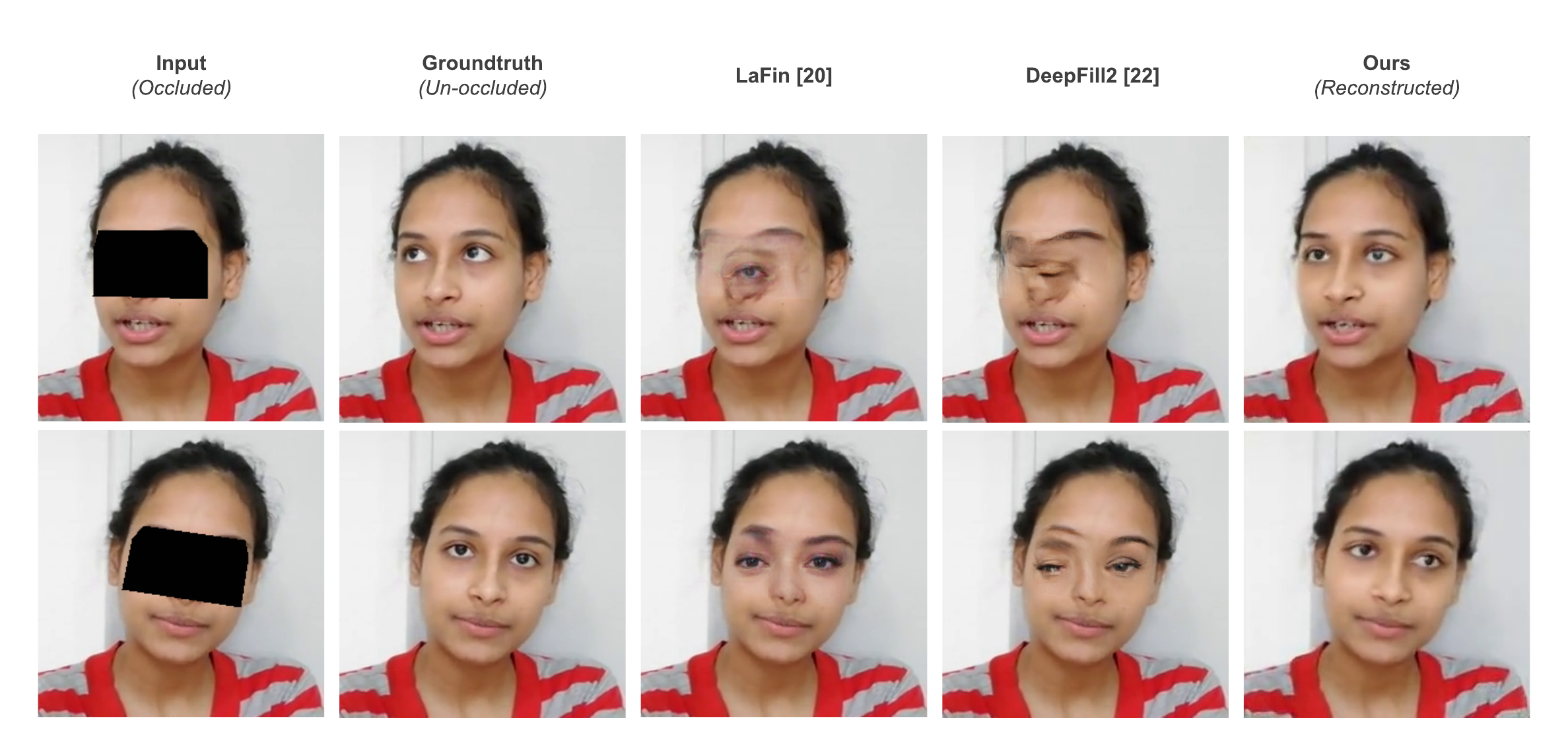}
    \caption{Comparison with state-of-the-art inpainting methods.}
    \vspace{0.615em}
    \label{fig:inpaint-comp}
\end{figure*}

Figure~\ref{fig:arch comparison} shows qualitative results comparing our method with baseline and variant of our method without mask loss. Here we use test examples that are completely unseen appearances (not part of the training set). We can observe here that the baseline model is unable to capture the background appearance as it tries to hallucinate it from the training examples, whereas introducing the attention framework allows the model to use the high-frequency background information from the input image. Furthermore, adding the mask loss introduces more consistency in the hallucination of the masked region. As can be observed in row 3 of Figure~\ref{fig:arch comparison}, introducing the mask allows the eyes to be open as it is with the ground truth face image. 
%
Similarly, Table~\ref{table:compare} reports the quantitative evaluation indicating the benefit of our approach for unseen appearances. We can observe that our method achieves superior results in terms of all evaluation metrics. It is important to note that though our method seems to perform only marginally better than LaFin~\cite{1} in terms of SSIM and LPIPS measures, yet there is a huge difference in terms of qualitative results where our method significantly outperforms LaFin~\cite{1} in terms of consistency of face appearance as shown Figure~\ref{fig:inpaint-comp}. This might be because, unlike our methods, their method only inpaint the mask region and hence might obtain a better score due to overlapping visible regions in input and output images. 

\begin{table}[!ht] 
\centering
\caption{Quantitative comparison with other methods on face reconstruction.}
\label{table:compare}
\begin{tabularx}{0.32\textwidth}{|l|c|c|c|}
\hline
Method & SSIM$\uparrow$ &  PSNR$\uparrow$ & LPIPS$\downarrow$\\ 
\hline
Baseline~\cite{7} & 0.706 & 19.627 & 0.176 \\
LaFin~\cite{1} & 0.914& 23.693 & 0.0601  \\
Ours  & 0.918 & 29.025 & 0.042\\ 
\hline 
\end{tabularx}
\end{table}

\begin{table}[!ht] 
\centering
\caption{Ablation Study on different loss functions.}
\label{table:ablation}
\begin{tabularx}{0.945\linewidth}{|l|c|c|c|}\hline
Method & SSIM$\uparrow$ &  PSNR$\uparrow$ & LPIPS$\downarrow$\\ 
\hline
Ours ($L_{rec}$)& 0.920 & 29.249 & 0.072 \\
Ours ($L_{rec}$ + $L_{adv}$)& 0.822 & 27.638 & 0.141  \\
Ours ($L_{rec}$ + $L_{adv}$ + $L_{ssim}$) & 0.916 & 28.973 & 0.045\\
Ours ($L_{rec}$ + $L_{adv}$ + $L_{ssim}$ +$L_{mask}$) & 0.918 & 29.025 & 0.042\\
\hline
\end{tabularx}
\end{table}
\begin{figure*}
\centering
    \includegraphics[width=\textwidth]{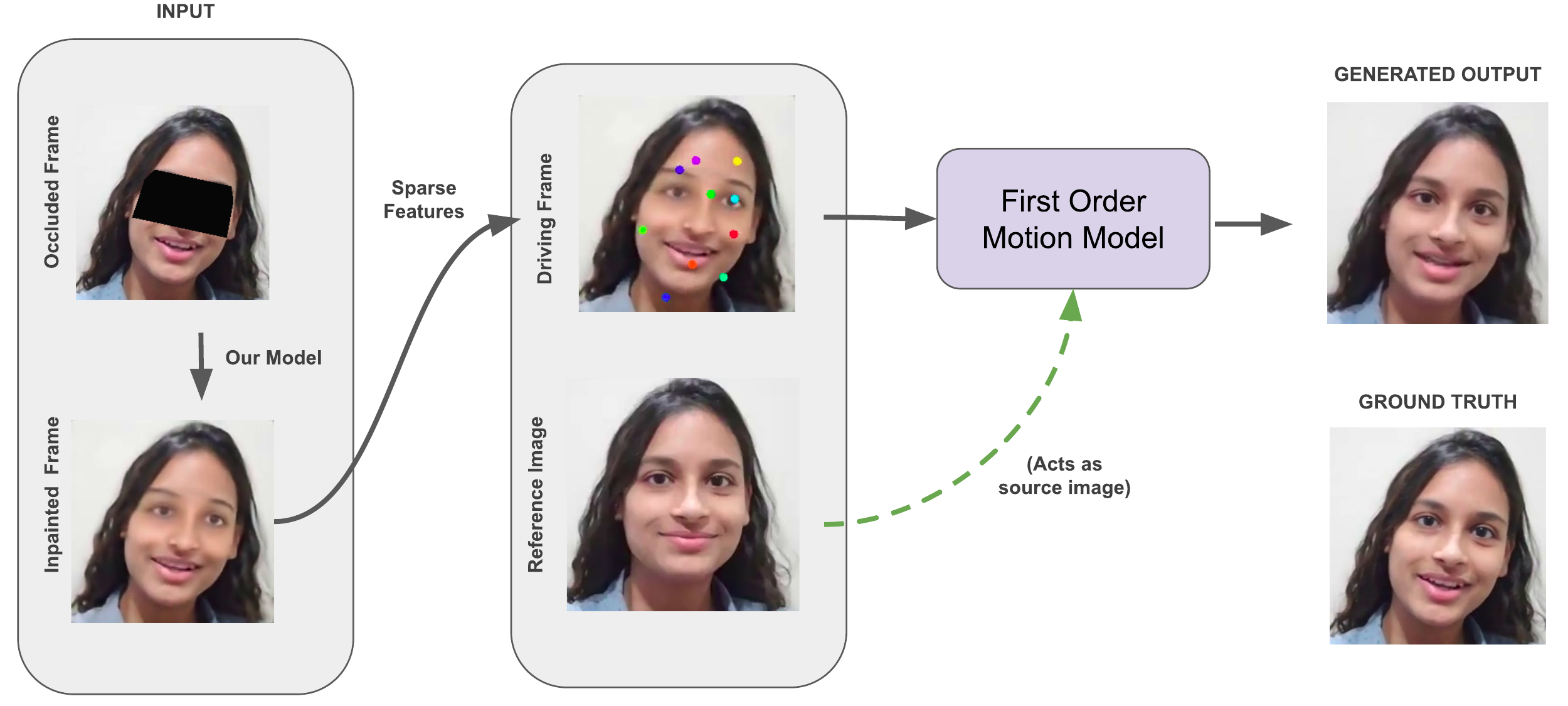}
    \caption{2D facial animation using FOMM\cite{5}.}
    \vspace{0.6em}
    \label{fig:fomm}
\end{figure*}
\subsubsection{Ablation Study on Losses}
\label{sec:ablation_study_on_losses}
We also perform an ablation study on the various loss functions used in our method and demonstrate their performance qualitatively and quantitatively. In Figure~\ref{fig:loss}, we show qualitative results on the unseen appearances after training with different loss functions. We can observe that training only with $L_{rec}$ generates a blurry image. The addition of $L_{adv}$ introduces more sharpness to the image. Finally, $L_{ssim}$  and $L_{mask}$ make it more consistent with facial features in the original image. Table~\ref{table:ablation} reports quantitative evaluation indicating incremental importance of all loss functions in our formulation. It is important to note that, although the reported SSIM and PSNR value of our combined loss function seems to be marginally lower than using only $L1$ loss, we can clearly observe the disparity in reconstructed images as shown in Figure~\ref{fig:loss}(F) where our combined loss yields superior reconstruction, including sharper facial features as well as consistent plausible reconstruction (closed eyes) as compared to using only $L1$ loss in as shown in Figure~\ref{fig:loss}(C).

\subsubsection{Qualitative comparison with SOTA Inpainting Methods}
\label{sec:quant_comparision_with_sota}
We also compute qualitative evaluation against some SOTA face inpainting methods. Figure~\ref{fig:inpaint-comp} demonstrates reconstruction results using our person-specific method on an unseen appearance v/s generalized inpainting from their pre-trained models. It can be observed that these methods fail to generalize well on images that are out of their training distribution and also suffer from severe loss of identity in frontal head poses itself. Nevertheless, we need to keep this in mind while comparing that our model is user-specific and has been trained on videos of the same user (but in different appearance), whereas their model is generic. 
Please refer to our supplementary video for an extended set of video results.

\balance
\section{Application to Hybrid Telepresence System}
\label{sec:hybrid-telepresence}
Recent works on face video animation, such as \cite{5},\cite{one-shot}, demonstrate that by just using sparse landmarks, a face image can be animated reasonably well from a reference image and show its application in low data bandwidth environments.

Figure~\ref{fig:fomm} shows how our approach can be easily integrated with these methods to handle HMD occlusions. We start with our inpainting module that can de-occlude the input image and reconstructs the missing region. Sparse features are then generated on top of this using~\cite{5} and used to drive a reference image. Thus, we can generate a consistent 2D video feed from the input occluded video feed. Additionally, this animated face can also be used for per-frame 3D face reconstruction tasks \cite{df2net} and fed to other VR teleconferencing users wearing a VR headset (as shown in Figure \ref{fig:fomm}). Thus, we can use our method to integrate users with or without VR headsets in a single hybrid teleconferencing application.
\begin{figure*}
\centering
    \includegraphics[width=0.95\textwidth]{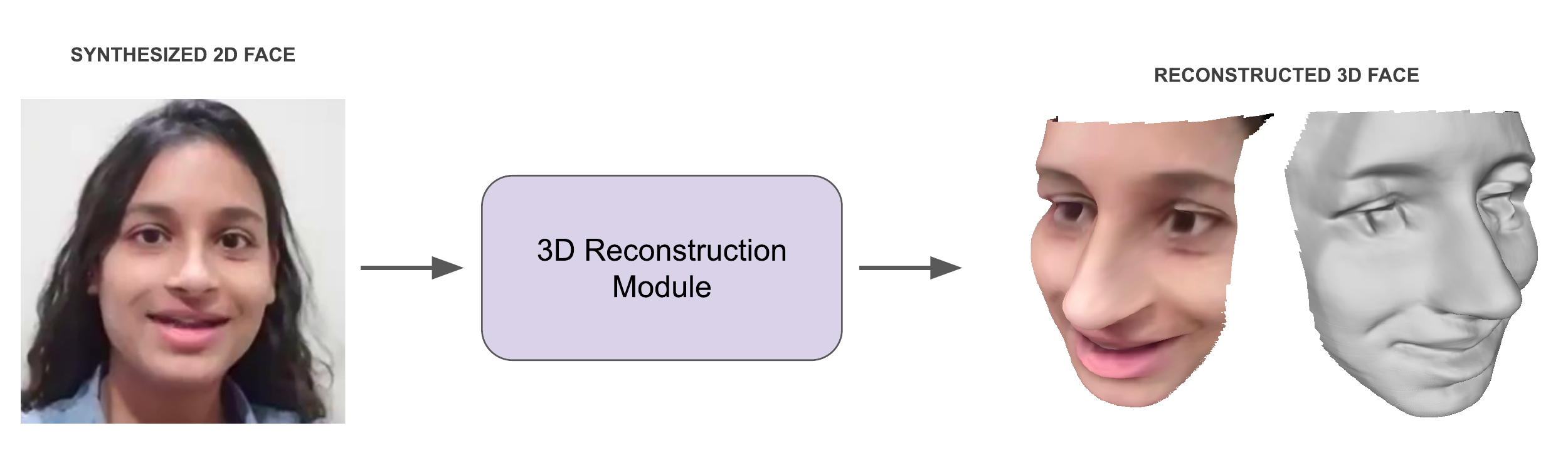}
    \label{fig:3drecon}
    \caption{ 3D reconstruction of de-occluded frame using DF2Net\cite{df2net}.}
\end{figure*}

\section{Discussion}
Our proposed approach promises to give superior results, both qualitatively and quantitatively, as shown earlier. However, there still exists a scope of improvement by incorporating eye tracking and gaze information for further refining our results. 
As future work, we would like to improve our work in facial video inpainting by leveraging temporal information across consecutive frames. Using modern HMD devices, we can give extra supervision about the eye information to the network to produces stable reconstruction in the eye region that is consistent across frames.

\section{Conclusions}

We proposed to learn a personalized model for face de-occlusion, particularly as an application in VR teleconferencing, where the face is partially occluded due to HMD. To tackle this, we formulate the face de-occlusion problem as an image inpainting task.
Our proposed attention enabled encoder-decoder network takes an HMD occluded face as input and completes missing facial features, particularly the eye region. The experiments show that our method works reasonably well with the same person wearing different clothes, facial appearances, poses, and expressions. To justify our proposed solution, we compared against a variety of related techniques such as image inpainting and facial animation and evaluated quantitatively and qualitatively. 

\balance
{\small
\bibliographystyle{ieee}
\bibliography{egbib}
}

\end{document}